# How well can LLMs Grade Essays in Arabic?

Rayed Ghazawi[1] and Edwin Simpson[2]

**Abstract**

This research assesses the effectiveness of state-of-the-art large language models (LLMs), including ChatGPT, Llama, Aya, Jais, and ACEGPT, in the task of Arabic automated essay scoring (AES) using the AR-AES dataset. It explores various evaluation methodologies, including zero-shot, few-shot in-context learning, and fine-tuning, and examines the influence of instruction-following capabilities through the inclusion of marking guidelines within the prompts. A mixed-language prompting strategy, integrating English prompts with Arabic content, was implemented to improve model comprehension and performance. Among the models tested, ACEGPT demonstrated the strongest performance across the dataset, achieving a Quadratic Weighted Kappa (QWK) of 0.67, but was outperformed by a smaller BERT-based model with a QWK of 0.88. The study identifies challenges faced by LLMs in processing Arabic, including tokenization complexities and higher computational demands. Performance variation across different courses underscores the need for adaptive models capable of handling diverse assessment formats and highlights the positive impact of effective prompt engineering on improving LLM outputs. To the best of our knowledge, this study is the first to empirically evaluate the performance of multiple generative Large Language Models (LLMs) on Arabic essays using authentic student data.

## 1 Introduction

Written assessments are a key method for evaluating students' performance, offering a deeper insight into their knowledge compared to other methods [1]. However, scoring essays is time-consuming and challenging for educators due to the significant effort required [1, 2]. AES alleviates this burden, allowing educators to focus more on task development and educational strategies while providing quicker feedback to enhance students' learning [3].

LLMs such as ChatGPT, Llama, Aya, Jais, and ACEGPT, have proven effective in various tasks including translation, summarization, text generation, and question-answering [4], raising the question of their potential for automated essay assessment. Unlike traditional models, LLMs can process longer documents and perform multi-step processing, apparently solving some reasoning tasks and following complex instructions. This adaptability allows them to be trained with minimal examples while still achieving effective results [5]. The ability to process instructions motivates the idea of passing essay marking criteria as part of a prompt, which for Arabic, has not previously been tested, or extensively in other languages. Recent investigations have examined the efficacy of LLMs in automatically scoring English-language essays, yielding promising outcomes [6], although their performance has yet to surpass state-of-the-art (SotA) benchmarks [7]. However, the evaluation of LLMs for AES in languages other than English, particularly Arabic, remains largely unexplored.

In response to this gap, the current research assesses the effectiveness of several prominent LLMs in the context of scoring Arabic essays. The study specifically evaluates ChatGPT, Llama, Aya, Jais, and ACEGPT, employing diverse methodologies such as zero-shot and few-shot learning, and fine-tuning. These evaluations are conducted utilizing the recently introduced AR-AES dataset of Arabic essay responses [8], comprising 12 essay prompts of varied types, including argumentative, descriptive, and source-based questions. The key contributions of this study are summarized as follows:

**Evaluation of LLMs for Arabic AES:** We compare several LLMs across different disciplines and find that they still do not outperform smaller pretrained models like BERT.

**Mixed-language Prompt Engineering Approach:** We found that mixing English prompts with Arabic content yielded performance gains of 49.49% compared to prompts given solely in Arabic or English, indicating the effectiveness of this bilingual strategy for Arabic AES.

**Addressing Challenges in LLMs for Arabic:** This study explores several challenges that LLMs encounter when processing Arabic, including tokenization complexities, which lead to increased computational demands, and strategies to mitigate these challenges. Additionally, the study examines the difficulties in crafting prompt engineering commands in Arabic and presents solutions to enhance the effectiveness of these prompts.

## 2 Related Work

Automated essay scoring (AES) represents a longstanding area of inquiry within natural language processing (NLP) [9]. While considerable attention has been directed towards the English language, evidenced by endeavors to engineer features that capture grammatical and lexical aspects of essays [10–12], subsequent studies have introduced neural networks [13] and hierarchical sentence-document models aimed at more comprehensive essay representation [14, 15]. Pretrained transformers, such as BERT, have achieved SotA results [16, 17], and have continued to outperform more recent LLMs, including ChatGPT and Llama2 [6, 7].

In contrast, research on AES in the Arabic language has not received comparable attention. Many studies in this area focus heavily on feature engineering, often relying on surface features that may not adequately capture the semantic and structural complexities of essays. These methods typically emphasize word-level or grammatical features, often neglecting aspects like word order. This limitation could be due to the linguistic intricacies of Arabic or the limited availability of dedicated Arabic datasets [18]. Various techniques have been explored in Arabic AES research, including linear regression [19], latent semantic analysis (LSA) [20], support vector machines (SVM) [21], rule-based systems [22], naïve Bayes [23] and optimization algorithms such as eJaya-NN [24]. Nevertheless, a recent study introduced a publicly-available dataset containing Arabic essays, paving the way for further research. Employing the pretrained model AraBERT, an Arabic variant of BERT, the results demonstrated promising performance [8]. Our current study is the first to explore larger, more recent LLMs for Arabic AES.

## 3 LLMs for Arabic AES

***Llama:*** We treat AES as an ordinal classification task, where the task is to select the correct grade for an essay [25]. Llama 2 is a multilingual model that has shown strong performance in various text classification tasks, such as binary sentiment classification in English, where it achieved an average accuracy of ∼91%, outperforming GPT-3.5 [26]. While the specific language proportions in Llama 2's training data have not been disclosed [27], it is clear that English text dominates, drawing from sources like CommonCrawl, C4, GitHub, Wikipedia, and ArXiv [28]. This emphasis on English suggests Llama 2 may perform better in English than other languages [29]. However, recent studies also indicate Llama's capability in handling Arabic [30], with Llama 2 reaching high accuracy in Arabic question-answering (93.70) [31]. In English-language AES, Llama 2 has proven effective in grading short answers and essays [32], supporting its use in our study for Arabic AES.

We employed two Llama models for fine-tuning: Llama 2 7B (7-billion parameters) for label-supervised adaptation and the OpenLLaMA model (3-billion parameters) for instruction fine-tuning [28]. OpenLLaMA's training incorporates a diverse range of datasets, including refined-web data from Falcon, the StarCoder dataset, and portions of Wikipedia, ArXiv, and StackExchange from RedPajama. Although explicit evidence of Arabic data inclusion is not provided, it is likely that Wikipedia entries in multiple languages, including Arabic, are part of the training data. Similarly, ArXiv, books, and StackExchange, though primarily in English, may contain some Arabic content.

***ChatGPT:*** ChatGPT has demonstrated effectiveness in a range of NLP tasks, including AES and Automatic Short Answer Grading (ASAG) in English [33–36]. Rasul et al. [37] examine the applications and potential of ChatGPT in educational settings, focusing on its capability to generate and evaluate long-form text, such as essays. The study emphasizes ChatGPT's effectiveness in following instructions and generating coherent, extended responses. In particular, ChatGPT-4 features a context window of up to 128,000 tokens. This enhancement enables the development of more extensive content, facilitates extended dialogues, and supports thorough document searches and analyses. The ability of ChatGPT to handle long texts is particularly beneficial for AES, which could involve lengthy marking criteria as part of the prompt and extended essays.

In terms of Arabic language capabilities, Ammar et al. [30] note that Arabic constitutes only 0.03% of GPT-3's training data, although ChatGPT-4 has shown a significant improvement in handling Arabic, with an 80% enhancement in Massive Multitask Language Understanding (MMLU) [38] over previous iterations [39]. Recent studies further indicate that ChatGPT-4 surpasses ChatGPT-3.5 in various metrics, demonstrating better precision and language comprehension, particularly in Arabic [40–42].

***Aya:*** An open-access multilingual LLM, trained on a diverse dataset including Arabic text sourced from Masader, which contains around 200 annotated datasets created after 2010 [43]. The data collection involved several repositories, including GitHub, Paperswithcode, Huggingface, LREC, Google Scholar, and LDC [44]. Aya outperforms multilingual models such as mT0 and BLOOMZ across a broader range of tasks and supports twice as many languages. However, its specific performance in AES remains to be thoroughly investigated. Aya's training data is approximately 2.5 times the size of xP3 dataset, which contains 81.2 million data points. xP3 is a widely recognized multilingual dataset designed for instruction-tuned models and serves as a benchmark for evaluating the scalability and multilingual capabilities of language models [45]. Aya's comparatively larger and more diverse dataset includes 101 languages, with only 21.5% of its data in English, compared to 39% in xP3.

***ACEGPT:*** Developed to address the limitations of generic LLMs in non-English contexts, ACEGPT particularly focuses on Arabic language proficiency and cultural sensitivity [46]. ACEGPT was pretrained on a large Arabic corpus, including datasets such as Arabic Text 2022, Arabic Wikipedia, CC100, and OSCAR. This extensive pretraining improves the model's understanding of Arabic grammar, vocabulary, and cultural nuances. ACEGPT also uses instruction-tuning data sets with questions translated from English and Arabic responses generated by GPT-4 (e.g., Alpaca, Evol-Instruct) and integrates reinforcement learning with an Arabic culture-aware reward model. Built on the Llama 2 architecture, ACEGPT achieves state-of-the-art performance among open-source Arabic LLMs across various benchmarks, including instruction following, natural language understanding, knowledge access, and cultural alignment. ACEGPT is available in multiple versions, including AceGPT-13B.

***Jais:*** Jais is an Arabic instruction-tuned generative LLM developed using a standard transformer-based decoder-only architecture, similar to GPT-2 and LLaMA [29]. The model is pretrained on a diverse corpus consisting of Arabic (29%), English (59%), and programming code (12%). The tokenizer used in Jais is trained on a balanced Arabic and English corpus using byte-pair encoding (BPE) with a dataset of 10 billion words. This approach mitigates bias towards either language and addresses the common tokenization issue in many LLMs, where text, particularly in Arabic, is often over-segmented into individual characters. Such over-segmentation not only reduces model performance but also increases computational costs.

Jais was trained on translated versions of English instruction-tuning datasets alongside Arabic-specific instruction datasets. The model is available in various sizes, ranging from 590 million to 70 billion parameters. Jais has demonstrated state-of-the-art performance in generative tasks, including machine translation [47]. However, its capabilities in classification tasks, particularly in Arabic, remain a subject of ongoing evaluation [30]. Further research is needed to comprehensively assess its effectiveness in Arabic AES.

# 4 The Experiments

## 4.1 The Dataset

We train and evaluate our models on the AR-AES dataset [8], containing 2,046 essays written by male and female undergraduate students at Umm Al-Qura University, representing a range of disciplines and writing styles. The essays cover three question types: argumentative, narrative, and source-dependent [48]. This inclusion allows for the evaluation of AES models in various categories of essays, which include persuasive writing, analysis of source materials, and storytelling – categories where LLMs could have an advantage over smaller models. The essays originate from both *traditional* in-person and *online* exams, reflecting the contemporary assessment landscape, as shown in Table 1.

The models were trained and evaluated using the course director's marks as ground truth. We conducted assessments on the entire dataset, as well as on individual courses and questions. This approach enabled us to explore two methods: a general-purpose model applicable to all courses and questions, and specialized models tailored for specific courses or questions, simulating the different ways that AES models could be trained in a real-world application. For fine-tuning, the dataset was split into training, validation, and test sets in a 70/15/15 ratio, following the methodology outlined in [8].

**Table 1**: Summary of the AR-AES dataset showing exam types (T=traditional, O=online) and range of answer lengths (number of tokens as measured by the AraBERT tokenizer), and the train/validation/test split sizes for each course [8].

| Course | Exam Type | Question ID | Answer Length Max | Answer Length Min | Split Size Train | Split Size Val | Split Size Test |
|---|---|---|---|---|---|---|---|
| Entire Dataset | | All questions | 575 | 2 | 1432 | 307 | 307 |
| Introduction to Info Science | T | All questions | 298 | 2 | 586 | 126 | 126 |
| | | Q1 (Narrative) | 298 | 7 | | | |
| | | Q2 (Argumentative) | 164 | 2 | 195 | 42 | 42 |
| | | Q3 (Source Dependent) | 61 | 4 | | | |
| Management Info Systems | T | All questions | 512 | 16 | 380 | 81 | 81 |
| | | Q4 (Narrative) | 512 | 29 | | | |
| | | Q5 (Narrative) | 212 | 29 | 127 | 27 | 27 |
| | | Q6 (Source Dependent) | 171 | 16 | | | |
| Environmental Chemistry | O | All questions | 422 | 8 | 244 | 52 | 52 |
| | | Q7 (Narrative) | 422 | 25 | | | |
| | | Q8 (Argumentative) | 116 | 9 | 81 | 17 | 17 |
| | | Q9 (Source Dependent) | 92 | 8 | | | |
| Biotechnology | O | All questions | 575 | 11 | 223 | 48 | 48 |
| | | Q10 (Source Dependent) | 357 | 13 | | | |
| | | Q11 (Argumentative) | 538 | 11 | 74 | 16 | 16 |
| | | Q12 (Source Dependent) | 575 | 13 | | | |

## 4.2 Tokenization

LLM tokenizers often encounter limitations when dealing with languages beyond English, including Arabic [30, 49]. In models like Llama and ChatGPT, tokenization splits each Arabic character into an individual token rather than treating words or phrases as single units [29]. This approach increases sequence length and can negatively impact model performance, as individual characters lack independent meaning. This results in higher GPU memory usage and potential performance degradation. For example, the phrase "مرحبا" (English "Hello") is tokenized into five separate tokens instead of a single word-level token, as shown in Figure 1, compared to just one token for its English equivalent.

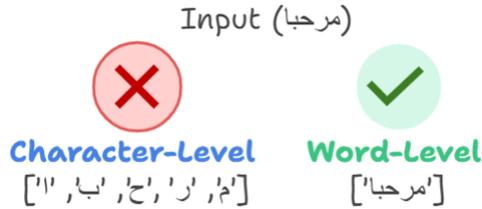

**Figure 1**: An example of tokenization challenges in LLMs for Arabic, where the text is tokenized at the character level rather than the word level. Illustrated here with the Arabic word "مرحبا" ("welcome").

This happens because most transformers employ wordpiece or byte pair encoding algorithms, which build a vocabulary based on the training set [28]. When trained on datasets containing little Arabic, Arabic words are not added to the vocabulary and so, when the tokenizer is applied, the unrecognized words are split into single characters [29]. This issue also causes the decoder to generate individual characters one-at-a-time, which may limit its ability to generate complex, multi-character words and sentences.

The need for tokenizer optimization, especially for languages with morphological and syntactic structures that differ significantly from English has been highlighted in previous research [50], which points out how tokenization discrepancies in LLMs create fairness challenges. Firstly, there's a cost disparity: users of certain languages may end up paying more than 2.5 times the amount English users pay for the same task, thanks to the increased sequence lengths. Secondly, there's the issue of latency: processing time can increase twofold for languages that require more tokens. Lastly, there's the challenge of long context processing: some language models can handle significantly longer texts than others, impacting service quality.

To specifically address these challenges with the Llama model, we developed a custom SentencePiece tokenizer trained on the AR-AES dataset (the same dataset used for evaluation) to capture Arabic words at the word level rather than the character level. To integrate the new Arabic tokens, we concatenated them with the existing Llama vocabulary, expanding the tokenizer's vocabulary to include both original and new tokens. This required an update to the embedding layer to accommodate the expanded vocabulary, ensuring the model could interpret the newly added Arabic tokens seamlessly alongside the original tokens. Furthermore, we fine-tuned the Llama model to enable it to process the newly introduced tokens. This enhanced tokenizer facilitates character-level processing and represents text at the word level, resulting in shorter sequences and reduced GPU memory usage.

A comparative evaluation of both tokenizers was conducted using Llama 2 for the Arabic AES task on a randomly selected subset of the dataset. The results showed a slight improvement in evaluation metrics, particularly in the Quadratic Weighted Kappa (QWK) (see definition in Section 5), when using the enhanced tokenizer (Figure 2). The main benefit of the improved tokenizer is its ability to perform experiments with lower computational demands and reduced GPU usage, reducing the average sequence length from 1532 to 410 tokens. A similar approach was then applied to the ACEGPT model, producing similar improvements.

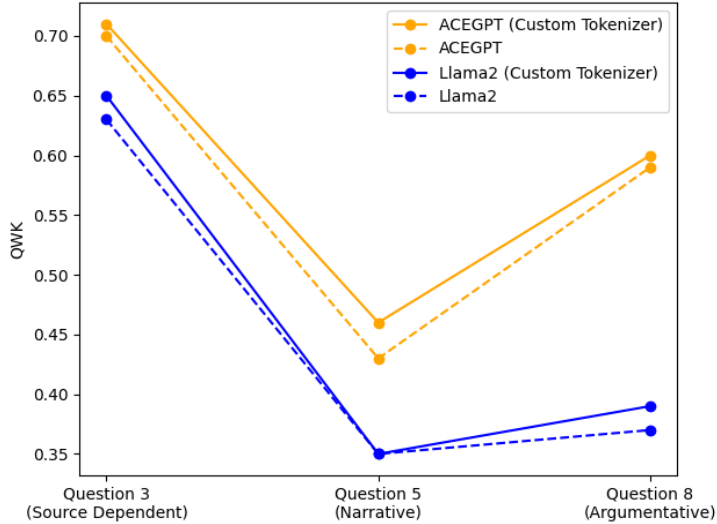

**Figure 2**: QWK scores for Llama2 and ACEGPT with standard and custom tokenizers across three question types, showing slight improvements with custom tokenization, highlighting the impact of Arabic-specific tokenization on performance.

### 4.3 Data Preprocessing

In the fine-tuning approach, we applied data preprocessing, while for few-shot and zero-shot tasks, the data was used in its original form. Preprocessing involved removing punctuation, hashtags, URLs, redundant letter repetitions, emoticons, extra spaces, numerical characters, and diacritics. These steps were taken to reduce noise in the text and focus the models on linguistic content rather than irrelevant features. This approach is consistent with findings by Kwon et al. [51], which demonstrate improved performance of fine-tuned LLMs in Arabic when punctuation is excluded, highlighting the LLMs' challenges in handling punctuation due to the lack of standardized punctuation rules in Arabic. Specific Arabic characters were normalized to standard forms (e.g., ة (tā) to ه (hā), ى (alif maqṣūrah) to ي (yā), ا,أ,إ,أ (different forms of alif) to ا standard alif , ئ, ؤ (hamzah) to stand-alone hamzah ء, before tokenization, as Arabic users often employ alternative spellings or character forms in informal contexts. To further simplify Arabic text and reduce vocabulary diversity, we applied the ISRI Stemmer, following Ghazawi and Simpson [8].

We conducted experiments with Llama2 and ACEGPT to evaluate model performance with and without preprocessing. Results indicated a minimal difference, with preprocessing yielding an average increase of 0.02 in Quadratic Weighted Kappa (QWK) for fine-tuned Llama and 0.01 for fine-tuned ACEGPT. Therefore, the following experiments fine-tuned the models with this stemmed and normalized text.

### 4.4 Training and Scoring Approaches

Three primary approaches were explored: zero-shot learning, few-shot in-context learning, and fine-tuning. Each approach offers distinct advantages that are particularly relevant to AES.

***Zero-shot learning:*** without any task-specific training examples, relying on model capabilities derived from pretraining allows us to apply only the marking guidelines and scheme without requiring specific examples. This approach is efficient and cost-effective as it removes the need for fine-tuning and does not require the course instructor to provide examples of essay answers. [52].

***Few-shot in-context learning:*** allows the model to learn from a small number of examples provided within the prompt. This does not involve updating the model's weights, avoiding the costs of fine-tuning and making it particularly useful in situations where gathering large amounts of labeled data is impractical. However, it may not be as effective as fine-tuned pretrained models [53].

***Fine-tuning:*** although requiring more human effort to provide more training examples than in-context learning, is highly effective with a relatively small number of examples, especially when task-specific data is available, and was shown to outperform in-context learning in short answer scoring [54]. In this study, we employed two fine-tuning methods: instruction tuning and label-supervised adaptation. Both approaches leveraged Low-Rank Adaptation (LoRA), a technique that updates only a small subset of the model's parameters. LoRA enables efficient fine-tuning, making it a practical choice for domain-specific applications such as Arabic Automated Essay Scoring (AES) [55].

***Instruction fine-tuning:*** tailors the model to follow specific prompts for essay scoring tasks, which may contain detailed instructions or examples. The model generates a numerical score as a response to these prompts, which we then extract from the output text using regular expressions, as illustrated in Figure 3. Loss is then computed by comparing the predicted label to the true label, then used to fine-tune the model's response. Each essay was paired with an instruction to assign a grade from 0 to 5. In this setup, the custom SentencePiece tokenizer, designed for Arabic, was used. Since LLMs were trained to generate human-like responses, instruction-tuning may not be ideal for tasks requiring fixed-label classification. Prior research indicates that instruction-tuned LLMs often underperform compared to BERT-based models in such classification tasks [56].

***Label-supervised adaptation (LS-LLaMA):*** the input text is processed by the LLM, generating latent representations from its final layer, which are then passed through a classification layer to compute class logits, with the final outputs being probabilities for each class, as shown in Figure 3. The cross entropy loss is then calculated between the predicted logits and the true labels, and used to fine-tune the model. This approach capitalizes on the strengths of LLMs and the information acquired during pretraining, while addressing the limitations of instruction tuning. Li et al. [56] observed that, even without intricate prompt engineering or external knowledge, LS-LLaMA significantly outperformed LLMs using instruction tuning that were ten times its size.

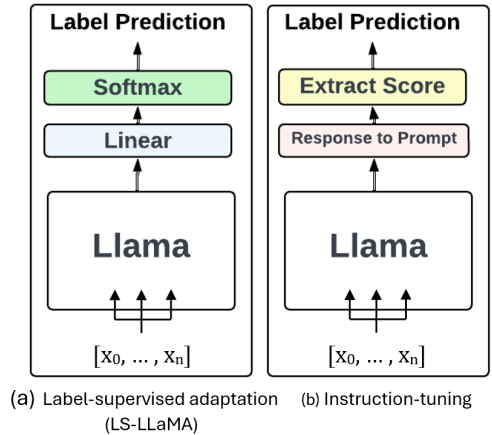

**Figure 3**: Comparison of label-supervised adaptation (LS-LLaMA) [56], and instruction tuning for Llama 2

## 4.5 Model Implementations

***Llama:*** For instruction-tuning we use the "Llama-2-7b-hf" model [27], and for LS-LLaMA we use Open-LLaMA 3B [57], which is based on the same pretrained model as Llama-2 but has fewer parameters. This makes it more suitable for tasks that require efficient fine-tuning and lighter computational resources. Additionally, we implemented a custom SentencePiece tokenizer specifically optimized for Arabic text to reduce tokenized sequence lengths and enhance processing efficiency.

***ACEGPT:*** We use LS-LLaMA with AceGPT-13B [46], using the same custom SentencePiece tokenizer. For few-shot in-context learning, we use the AceGPT-13B-chat model but retain the original model tokenizer.

***ChatGPT:*** ChatGPT-4 Turbo was used for zero-shot and few-shot learning tasks. The model is capable of handling complex tasks and processing longer text inputs compared to the other LLM models. In the few-shot learning configuration, the model was provided with three examples for each class to guide its predictions.

***Aya:*** This study applies few-shot in-context learning to the "Aya-101" model [43], a 13-billion-parameter autoregressive Transformer-based multilingual language model.

***Jais:*** This study used two configurations of the Jais model, the pretrained "Jais-family-13b" and the fine-tuned "Jais-family-13b-chat", both featuring a context length of 2048 tokens and 13 billion parameters [29]. The pretrained "Jais-family-13b" model was trained on a diverse corpus of Arabic and English texts, as well as source code, enabling broad language understanding. However, it was not specifically fine-tuned for particular tasks or conversational applications. In contrast, the fine-tuned "Jais-family-13b-chat" model builds upon the pretrained version and undergoes additional fine-tuning on curated datasets comprising prompt-response pairs in Arabic and English. This additional training aims to enhance the model's ability to handle task-specific and conversational contexts effectively. The models were evaluated in zero-shot and few-shot learning scenarios.

***AraBERT:*** The study introducing the dataset employed the large AraBERT model, comprising 12 encoder blocks, 1024 hidden dimensions, 16 attention heads, and a maximum sequence length of 512. The model has 370 million parameters. A classification head with a single fully connected layer was added to AraBERT for essay scoring [8].

## 5 Evaluation Metrics

We utilized Quadratic Weighted Kappa (QWK) as our primary evaluation metric. QWK, an extension of Cohen's kappa, measures the agreement level between the outcomes assessed by two raters. It is particularly favored in Automated Essay Scoring (AES) evaluations because it accounts for agreements that could occur by chance, providing a more reliable measure of scoring consistency compared to mere accuracy and F1 score. Furthermore, QWK is appropriate for essay assessments as it respects the ordinal nature of grading and incorporates quadratic weights to emphasize the significance of class rankings—a subtlety not captured by simple accuracy or F1 scores. While the AES models may be trained as nominal classifiers, it is necessary to evaluate their predictions as ordinal values, since larger errors in grade predictions would have a bigger negative impact.

The calculation of QWK involves a weighting factor defined as:

$$QWK = 1 - \frac{\sum_{i,j} w_{i,j} O_{i,j}}{\sum_{i,j} w_{i,j} n_{i,1} n_{j,2}}, \qquad (1)$$

where $w_{i,j} = \frac{(i-j)^2}{(N-1)^2}$ is the weight between mark $i$ and mark $j$, $N$ is the number of marks available, $O_{i,j}$ is the number of observations where the first assessor gave mark $i$ and the second assessor gave mark $j$, and $n_{i,k}$ is the number of times that assessor $k$ gave mark $i$.

# 6 Prompt Engineering

***ChatGPT:*** In the initial zero-shot experiments, we tested simple prompts in Arabic, such as "Correct the following essay answers out of 5 marks." However, this simplicity led to inconsistent performance, with the model providing varied outcomes, including unrequested feedback, or assigning scores on a 100% scale, rather than the expected five-point scale. To improve precision, we revised the prompts to be more specific: "Evaluate the essay by assigning one of the following grades: 0, 1, 2, 3, 4, or 5." While this generally produced valid grades, the model occasionally provided explanations or score ranges instead of a single score, which, although potentially useful to an essay marker, deviated from the primary objective of obtaining a final score for comparison with the actual score. To address these issues, we revised the prompts to provide clearer instructions: "Evaluate the following essay answers by assigning one of the six grades only (0-1-2-3-4-5), without any explanation or comment, only the final grade number." Despite improved results, additional text sometimes appeared with the score, (e.g., "The final grade is: 3 out of 5"), which was handled using regular expresions to extract the final grade.

In addition, we compared different combinations of Arabic text with English prompts on three questions chosen randomly from the narrative, source-dependent and argumentative categories (see Table 2). We observed that using English prompts instead of Arabic led to a noticeable increase in performance, improving the results by approximately 6%. Subsequently, the prompt was expanded to include detailed instructions in English, specifying the task, question, evaluation criteria (rubrics), golden answer, and evaluation scale, with improvement in the results on average. While this approach yielded better results, optimal performance was observed when instructions, such as the criteria for evaluation and the task description, were provided in English, while the essays, details of the question, rubrics, and standard answers remained in Arabic, improving the results by approximately 49.49%. This bilingual strategy avoided translating the Arabic essay answers into English, ensuring that the models evaluated the responses in their original form while leveraging the clarity of English for instructional prompts. This approach is illustrated in the last row of the zero shot section in Table 2, which highlights the improvements in accuracy under these conditions.

Building on this, a few-shot learning configuration was implemented, using the same detailed prompts with the addition of three example answers for each class. The "golden answer" served as a reference for the highest score, while the other examples represented varying levels of alignment with the ideal response, corresponding to different scores. This bilingual approach again surpassed a fully English few-shot setup, as shown by the results in Table 2.

Table 2: Zero-shot and few-shot prompt engineering experiments for three samples of different essay types (Q1, Q9, and Q11) using the ChatGPT-4 model.

| A | Language | Example of the Prompt | Q1 | Q9 | Q11 |
|---|---|---|---|---|---|
| Zero Shot | Arabic | صحح الإجابات المقالية التالية من 5 درجات (Mark the following essay out of 5 marks) | colspan: *Results not comparable* | | |
| | Arabic | قم بتقييم الإجابات المقالية التالية من خلال تعيين إحدى الدرجات الستة التالية فقط: 5,4,3,2,1,0 (Evaluate the following essay answers by assigning only one of the following six marks: 0,1,2,3,4,5) | 0.18 | 0.24 | 0.25 |
| | Arabic | قم بتقييم الإجابات المقالية التالية من خلال تحديد إحدى الدرجات الستة فقط (5,4,3,2,1,0)، بدون أي شرح أو تعليق، فقط العلامة النهائية (Evaluate the following essay answers by selecting only one of the following grades (0,1,2,3,4,5), without any explanation or comment, only the final mark) | 0.21 | 0.29 | 0.27 |
| | English | Evaluate the following essay answers by assigning one of the six grades only (0-1-2-3-4-5), without any explanation or comment, only the final grade number. | 0.23 | 0.30 | 0.28 |
| | English | **The Question:** What is the scientific definition of environmental chemistry?<br>**Criteria for Evaluation:**<br>- The student's ability to define the term scientifically (2.0 marks).<br>- The student's ability to identify the aspects addressed by environmental.. (2.0 marks).<br>- The student's ability to mention the importance of environmental chemistry for life... (1.0 mark).<br>**Standard Answer:** Environmental chemistry is the scientific study of chemical...<br>**Evaluation Scale:**<br>0 - Very poor. 1 - Poor. 2 - Fair. 3 - Good. 4 - Very good. 5 - Excellent.<br><br>**The Task:** Please mark the following essay answer based on the question, criteria for evaluation, standard answer, and evaluation scale. Select one of the six scores (0,1,2,3,4,5). Provide only the final score number without any additional comments or explanations. | 0.24 | 0.41 | 0.32 |
| | English & Arabic | **The Question:** ما هو التعريف العلمي لكيمياء البيئة؟<br>**Criteria for Evaluation:**<br>- قدرة الطالب على تعريف المصطلح بشكل علمي (0.2 درجة)<br>- قدرة الطالب على تحديد الجوانب التي يهتم بها علم الكيمياء البيئية (0.2 درجة)<br>- قدرة الطالب على ذكر أهمية كيمياء البيئة للحياة والإنسان (0.1 درجة)<br>**Standard Answer:**<br>كيمياء البيئة: هي الدراسة العلمية للظواهر الكيميائية والبيوكيميائية التي تحدث في الأماكن الطبيعية...<br>**Evaluation Scale:**<br>0 - Very poor. 1 - Poor. 2 - Fair. 3 - Good. 4 - Very good. 5 - Excellent.<br><br>**The Task:** Please mark the following essay answer based on the question, criteria for Evaluation, standard answer, and evaluation scale. Select one of the Six scores (0,1,2,3,4,5). Provide only the final score number without any additional comments or explanations. | 0.24 | 0.66 | 0.60 |
| Few Shot | English | **The Question:** What is the scientific definition of environmental ...?<br>**Criteria for Evaluation:**<br>- The student's ability to define the term scientifically (2.0 marks)<br>- The student's ability to identify the aspects in which... (2.0 marks)<br>- The student's ability to state the importance of... (1.0 marks)<br>**Standard Answer:** Environmental Chemistry: It is the scientific study of chemical and biochemical phenomena that occur in natural...<br>**Examples of Each Score:** (Three examples of each class)<br>Class (0) - Are the interactions resulting in the environment due to...<br>Class (1) - Study of land, air, water, living environments, and the...<br>Class (2) - Environmental Chemistry is the scientific study of...<br>Class (3) - It is the study of the sources, reactions, transfers, effects....<br>Class (4) - It is a science that specializes in studying the sources...<br>Class (5) - is the study of chemical processes that occur in water...<br>**Evaluation Scale:**<br>0 - Very poor. 1 - Poor. 2 - Fair. 3 - Good. 4 - Very good. 5 - Excellent | 0.24 | 0.56 | 0.61 |
| | English & Arabic | **The Question:** ما هو التعريف العلمي لكيمياء البيئة؟<br>**Criteria for Evaluation:**<br>- قدرة الطالب على تعريف المصطلح بشكل علمي (0.2 درجة)<br>- قدرة الطالب على تحديد الجوانب التي يهتم بها علم الكيمياء البيئية (0.2 درجة)<br>- قدرة الطالب على ذكر أهمية كيمياء البيئة للحياة والإنسان (0.1 درجة)<br>**Standard Answer:**<br>كيمياء البيئة: هي الدراسة العلمية للظواهر الكيميائية والبيوكيميائية التي تحدث في الأماكن الطبيعية...<br>**Examples of Each Score:** (Three examples of each class)<br>Class (0) هي التفاعلات الناتجة في البيئة بسبب تدخلات طبيعية أو بشرية مثل: البناء الضوئي ...<br>Class (1) دراسة الأرض والهواء والماء والبيئات المعيشية وآثار التكنولوجيا في هذا الشأن...<br>Class (2) هي الدراسة العلمية للظواهر الكيميائية التي تؤثر سلباً على البيئة مثال: التلوث...<br>Class (3) هي دراسة مصادر التفاعلات والتأثيرات الكيميائية في الماء...<br>Class (4) هو علم يختص بدراسة مصادر المواد الكيميائية وتفاعلاتها وتنقلاتها وتأثيراتها ومراقبتها...<br>Class (5) هي دراسة العمليات الكيميائية التي تحدث في المياه، الهواء، والبيئات الأرضية...<br>**Evaluation Scale:**<br>0 - Very poor. 1 - Poor. 2 - Fair. 3 - Good. 4 - Very good. 5 - Excellent<br><br>**The Task:** Please mark the following essay answer based on the question, criteria for Evaluation, standard answer, and evaluation scale. Select one of the Six scores (0,1,2,3,4,5). Provide only the final score number without any additional comments or explanations. | 0.25 | 0.74 | 0.64 |

***Aya:*** The Aya model was evaluated using a few-shot in-context learning approach with the same detailed prompt structure employed for ChatGPT-4. These prompts incorporated the question, the gold (standard) answer, evaluation criteria (rubrics) for each question, and the evaluation scale. The model exhibited an effective ability to follow structured prompts, including bilingual prompts, and most of the time generated only the final score as specified by the instructions.

***ACEGPT:*** We evaluated the ACEGPT model in a few-shot in-context learning setting, using the same bilingual, fully instructed prompt structure previously applied to ChatGPT-4. The model adhered closely to the given prompts and typically returned only the final score as instructed (e.g., "*The final score is: 1*").

***Jais:*** The same prompts used for ChatGPT were applied to Jais in both zero-shot and few-shot learning scenarios. In the few-shot experiments conducted for the entire dataset and at the course level, the number of examples per class was reduced to one instead of three to address the context window limitation of Jais, which supports up to 2048 tokens. Furthermore, the Jais chat model required specific command structures to delineate the roles of the user and the model. These structures included "`### Instruction:`" to introduce the prompt and model role, and "`### Input:[Human|]|`" to precede the task and the provided essay.

***Llama instruction experimentation:*** We carried out prompt engineering in two phases for Llama instruction tuning. Initially, during the preparation of the training data, we created an instruction dataset for each essay, including task instructions, essay content, and the actual score. In the second phase, we assessed the model's performance using various prompt configurations, including full Arabic and English prompts. The highest performance was achieved with prompts using English instructions and Arabic content, resulting in a QWK of 0.51 on the entire dataset (Figure B1). Full English prompts scored a QWK of 0.46, while full Arabic prompts performed the worst with a QWK of 0.27, indicating that Llama may not be fully adapted to the Arabic language.

# 7 Results and Discussion

Our study used the experimental protocol delineated by Ghazawi and Simpson [8]. We adapted this framework for our exploratory analysis involving four LLMs (Llama, ChatGPT, Aya, Jais, and ACEGPT), initially evaluating the models across the entirety of the dataset. This preliminary evaluation aimed to gauge the models' performance when exposed to a wide variety of data and diverse question types. Subsequently, we embarked on a more granular analysis, where each model was trained and evaluated separately on the data of individual courses. Following the course-specific evaluations, the analysis was further refined by assessing each model's performance on distinct questions, allowing us to examine performance with more specialized fine-tuning, albeit with fewer examples. Additionally, we analyzed the models' performance based on essay types (argumentative, narrative, and source-dependent). This multi-tiered approach ensured a comprehensive understanding of each model's strengths and weaknesses in diverse educational and assessment contexts.

***Overview of results:*** As shown in Table 3, the ACEGPT model using a fine-tuning approach achieved the highest QWK score among LLMs (excluding AraBERT), of 0.67 across the entire dataset, indicating strong alignment with human scoring. In the few-shot setting, ACEGPT outperformed other Arabic LLMs (Jais and Aya) also outperformed Llama 2 in instruction tuning, although it did not surpass ChatGPT 4, possibly due to ChatGPT 4's extensive training data. Conversely, Aya's few-shot approach yielded the lowest performance, with a QWK of only 0.13. The success of ACEGPT likely stems from its substantial Arabic-language pretraining and the advantages provided by the LLaMA-based architecture.

At the course level, the effectiveness of each model varied across courses. For instance, ChatGPT-4 using few-shot learning excelled in the "Introduction to Information Science" course and outperformed other fine tuning models, while Llama (LS-LLaMA) performed best in the other three courses. At the question-specific level, ChatGPT-4 demonstrated satisfactory performance, particularly in zero-shot learning. However, Llama and ACEGPT generally outperformed the other models, due to the advantages of fine-tuning over few-shot and zero-shot methods, particularly in handling Arabic. This highlights the variation in how models adapt to different question formats and complexities (Figure 4).

For essay types, ACEGPT (fine tuning), Aya, and ChatGPT-4 performed best in argumentative essays. Argumentative essays involve complex reasoning, and although the models performed reasonably well, there is room for improvement in consistently handling such challenging essay types. In contrast, Llama (LS-LLaMA), ACEGPT (few-shot), and Jais achieved their highest performance on source-dependent essays, which involve analyzing and synthesizing information from provided materials. This distinction highlights the varying strengths of the models across different essay types.

**Table 3**: Comparative performance (QWK) across models and training approaches.

| Approach | Fine-Tuning | | | | Few-shot | | | | Zero-shot | |
|---|---|---|---|---|---|---|---|---|---|---|
| Model/ Dataset | AraBERT | Llama2 (LS-LLaMA) | Llama2 (Instruction) | ACEGPT | ACEGPT | Jais | ChatGPT4 | Aya | Jais | ChatGPT4 |
| **Entire Dataset** | *0.88* | 0.64 | 0.51 | **0.67** | 0.55 | 0.20 | 0.64 | 0.13 | 0.14 | 0.45 |
| **Introduction Info Science** | *0.79* | 0.57 | 0.06 | 0.56 | 0.53 | 0.36 | **0.58** | 0.34 | 0.04 | 0.53 |
| **Management Info Systems** | *0.78* | **0.53** | 0.02 | 0.51 | 0.21 | 0.19 | 0.41 | 0.23 | 0.02 | 0.38 |
| **Environmental Chemistry** | *0.97* | **0.85** | 0.09 | 0.85 | 0.54 | 0.29 | 0.74 | 0.41 | 0.03 | 0.57 |
| **Biotechnology** | *0.95* | **0.83** | 0.06 | 0.74 | 0.27 | 0.28 | 0.35 | 0.20 | 0.27 | 0.33 |
| **Question 1** | *0.89* | **0.59** | 0.01 | 0.55 | 0.32 | 0.17 | 0.25 | 0.09 | 0.07 | 0.24 |
| **Question 2** | 0.73 | 0.61 | 0.25 | **0.78** | 0.34 | 0.30 | 0.43 | 0.21 | 0.60 | 0.52 |
| **Question 3** | *0.87* | 0.65 | 0.16 | 0.71 | 0.69 | 0.46 | 0.71 | 0.53 | 0.01 | **0.75** |
| **Question 4** | *0.83* | **0.60** | 0.19 | 0.52 | 0.18 | 0.27 | 0.33 | 0.06 | 0.16 | 0.51 |
| **Question 5** | *0.84* | 0.35 | 0.41 | 0.46 | 0.21 | 0.38 | **0.81** | 0.09 | 0.12 | 0.51 |
| **Question 6** | *0.94* | **0.48** | 0.01 | 0.41 | 0.14 | 0.10 | 0.09 | 0.09 | 0.04 | 0.31 |
| **Question 7** | *0.43* | **0.77** | 0.11 | 0.70 | 0.21 | 0.30 | 0.61 | 0.01 | 0.23 | 0.65 |
| **Question 8** | *0.79* | 0.39 | 0.17 | 0.60 | 0.25 | 0.29 | **0.93** | 0.34 | 0.15 | 0.61 |
| **Question 9** | *0.98* | 0.79 | 0.18 | **0.79** | 0.59 | 0.48 | 0.74 | 0.44 | 0.37 | 0.66 |
| **Question 10** | *0.90* | **0.59** | 0.37 | 0.58 | 0.55 | 0.07 | 0.47 | 0.35 | 0.13 | 0.68 |
| **Question 11** | *0.84* | 0.67 | 0.13 | **0.67** | 0.42 | 0.18 | 0.64 | 0.38 | 0.40 | 0.60 |
| **Question 12** | 0.84 | 0.46 | 0.06 | **0.81** | 0.36 | 0.31 | 0.50 | 0.20 | 0.25 | 0.65 |
| **Source Dependent (Q3,Q6,Q9, Q10,Q12)** | *0.89* | 0.59 | 0.15 | **0.66** | 0.47 | 0.28 | 0.50 | 0.31 | 0.16 | 0.61 |
| **Argumentative (Q2,Q8,Q11)** | *0.73* | 0.56 | 0.18 | **0.68** | 0.34 | 0.26 | 0.67 | 0.31 | 0.38 | 0.58 |
| **Narrative (Q1, Q4,Q5, Q7)** | *0.69* | **0.58** | 0.18 | 0.56 | 0.23 | 0.28 | 0.47 | 0.08 | 0.15 | 0.48 |

AraBERT consistently outperforms the four LLMs used in this study. It is notable that, unlike the LLMs, AraBERT is stronger with source-dependent essays than with argumentative essays. This can be attributed to its comprehensive training on diverse Arabic datasets and the effectiveness of its bidirectional encoder architecture in capturing the information needed for assigning a class label to a text sequence. AraBERT's strong performance reinforces the importance of specialised training for language models tailored to specific linguistic contexts. These findings align with previous research, such as Kocoń et al. [58], showing that while models such as ChatGPT are versatile, they may not match the performance of fine-tuned models like BERT in specialised tasks when only small amounts of training data are available [56]. This further supports our conclusion that AraBERT's fine-tuning on Arabic data allows it to excel in Arabic language processing compared to more general LLMs.

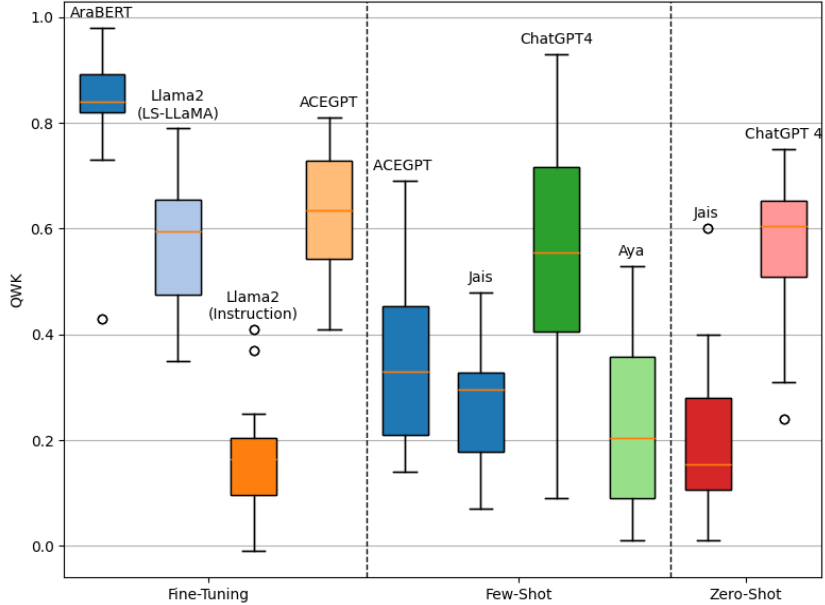

**Figure 4**: Ranges of QWK scores on the question-level across different models using fine-tuning, few-shot, and zero-shot approaches.

*Learning Approaches:* Zero-shot learning with Jais and ChatGPT-4 yielded suboptimal results, particularly for Jais, which achieved its highest QWK of 0.60 on Question 2. ChatGPT-4 performed better, with a QWK of 0.75 on Question 3, but zero-shot learning remained inadequate for the AES task compared to fine-tuned Arabic models. Few-shot learning with Aya demonstrated relatively poor performance, followed by Jais, which showed slightly better results. In contrast, ChatGPT-4 exhibited notable improvement in the few-shot setting, achieving a QWK of 0.93 on Question 8, surpassing AraBERT's performance (QWK 0.79). ACEGPT achieved results close to ChatGPT and outperformed Jais and Aya in most cases. Despite this improvement, fine-tuned models generally outperformed those using few-shot learning across the dataset (Figure 4). Instruction tuning with Llama 2 produced poor results due to challenges in Arabic prompt engineering. In contrast, label-supervised adaptation (LS-LLaMA) performed best, particularly with the ACEGPT model, followed by Llama 2, proving more effective for Arabic AES tasks by eliminating the need for prompt engineering.

In summary, zero-shot and few-shot in-context learning approaches underperformed compared to fine-tuned models, largely due to their reliance on pretrained knowledge without task-specific LS-LLaMA fine-tuning, which limits their capacity to assess essays according to specific criteria. Furthermore, the training data for these models may not adequately capture the nuances of Arabic essay scoring, in contrast to the specialized data used for fine-tuned models such as AraBERT. Instruction tuning with Llama 2 faced similar challenges in Arabic prompt engineering, while LS-LLaMA showed superior performance by handling Arabic AES tasks more effectively and reducing the need for prompt engineering, making it a more adaptable solution.

***Arabic vs. General-Purpose Models:*** The results suggest that ACEGPT, an Arabic-focused model, generally outperforms Llama2, a more general-purpose model, particularly in argumentative essays. However, comparisons to ChatGPT-4, Jais, and Aya are complicated by their different training objectives and methodologies. In few-shot settings, ACEGPT also demonstrated strong performance, surpassing Llama2's instruction fine-tuning and outperforming other Arabic models such as Jais and Aya. Although Jais was trained on a substantial Arabic corpus, it struggled with AES tasks and frequently did not adhere to instructions or prompts, possibly because it was developed as a generative model rather than a classification-focused one [30]. Nonetheless, the chat-based version of Jais showed improved compliance compared to its pre-trained form.

Although the Aya model is multilingual, it performed comparatively poorly in this task. By design, Aya follows the mT5 architecture for text-to-text transformations and is intended to handle multiple languages [43], which may not optimize performance for Arabic alone—particularly if the developer did not extensively adapt it for Arabic-specific linguistic features. In contrast, AceGPT builds on Llama2 and is fine-tuned for generative Arabic tasks [46], incorporating additional Arabic-centric training that better captures the linguistic nuances required for effective essay scoring.

Overall, some Arabic-specific models, such as ACEGPT, have performed better than certain general-purpose models (e.g., Llama) in Arabic essay scoring when fine-tuned for classification tasks. However, not all Arabic-centric systems excelled; Aya and Jais, despite their pretraining focus on Arabic, did not outperform general-purpose models (e.g., ChatGPT) with the same evaluation setup. In addition, limited information regarding ChatGPT's Arabic data volume, along with archeticture differences, makes direct comparisons difficult. Still, a comparison of ACEGPT and Llama within the same architecture and with increased Arabic data indicates that ACEGPT achieves stronger performance in these settings.

***Accuracy of Predictions within One Grade:*** We investigate the scale of the LLMs' errors, as differences of one grade from the human marker are less serious errors that could be caused by borderline essays. Therefore, we quantified the cases where the predicted score was within one grade of the actual score. In the zero-shot approach, Jais predicted within one grade 49.24% of the time across the entire dataset. ChatGPT-4 achieved higher accuracy at 72.06%, with its best performance on Question 4, where it reached 95.83% prediction accuracy (Table A1). In the few-shot approach, ChatGPT-4 outperformed both the zero-shot and fine-tuning methods. For instance, in the Management Information Systems course, the model predicted within one score of the actual score in 91.30% of cases, and showing the highest accuracy across all questions. These results highlight the models' ability to make reliable predictions in close proximity to actual scores.

***Interrater Agreement*** As reported by [8] and shown in Table 4, the highest agreement between two human expert raters was observed in Question 9, with a Quadratic Weighted Kappa (QWK) of 0.90. Notably, ChatGPT-4 in few-shot learning achieved a QWK of 0.93 on Question 8, outperforming the human raters, who achieved a QWK of 0.78 on the same question. Furthermore, the Large Language Models (LLMs) demonstrated high agreement with the course director's marks, surpassing human agreement in most of the questions (8 out of 12).

Higher human agreement was predominantly observed in source-dependent question types (Questions 6, 9, and 10), which are easier to score objectively by directly comparing student answers with course materials. In contrast, argumentative and narrative essay types require more subjectivity in marking. The Llama2 (LS-LLaMA) model achieved high agreement on narrative question types, which are more complex and require holistic evaluation of the student's response. Similarly, ChatGPT-4 in few-shot learning demonstrated high agreement on both argumentative and narrative essay types. These results indicate that LLMs have a strong capability to match or exceed human agreement levels when provided with clear prompts that include the necessary details for the automatic essay scoring task, even with complex essay types.

**Table 4**: Inter-rater Agreement Compared with Best Performing LLM Models in Different Learning Settings (Fine-tuning, Few-shot, and Zero-shot)

| Question ID | Question Type | Human | Fine-Tuning | | Few-shot | Zero-shot |
|---|---|---|---|---|---|---|
| | | | Llama2 | ACEGPT | ChatGPT4 | ChatGPT4 |
| Question 1 | Narrative | 0.52 | 0.59 | 0.55 | 0.25 | 0.24 |
| Question 2 | Argumentative | 0.56 | 0.61 | 0.78 | 0.43 | 0.52 |
| Question 3 | Source dependent | 0.68 | 0.65 | 0.71 | 0.71 | 0.75 |
| Question 4 | Narrative | 0.51 | 0.60 | 0.52 | 0.33 | 0.51 |
| Question 5 | Narrative | 0.79 | 0.35 | 0.46 | 0.81 | 0.51 |
| Question 6 | Source dependent | 0.66 | **0.48** | 0.41 | 0.09 | 0.31 |
| Question 7 | Narrative | 0.75 | 0.77 | 0.70 | 0.61 | 0.65 |
| Question 8 | Argumentative | 0.78 | 0.39 | 0.60 | 0.93 | 0.61 |
| Question 9 | Source dependent | 0.90 | 0.79 | **0.79** | 0.74 | 0.66 |
| Question 10 | source dependent | 0.83 | 0.59 | 0.58 | 0.47 | **0.68** |
| Question 11 | Argumentative | 0.78 | 0.67 | **0.67** | 0.64 | 0.60 |
| Question 12 | source dependent | 0.68 | 0.46 | 0.81 | 0.50 | 0.65 |

# 8 Conclusion and Future Works

This study evaluated a range of prominent large language models (LLMs), ChatGPT4, Llama 2, Aya, Jais, and ACEGPT, for automatic scoring of Arabic essays in the AR-AES dataset. We explored zero-shot, few-shot, and fine-tuning approaches. A key preprocessing challenge was tokenization, which we addressed by developing an optimized SentencePiece tokenizer to reduce the length of tokenized Arabic sequences, thereby decreasing memory usage and computational overhead. ACEGPT showed the strongest overall performance among LLMs, with a QWK score of 0.67, reflecting its extensive training on Arabic-specific data, which allowed it to consistently outperform the general-purpose Llama model. Fine-tuned models, particularly those using label-supervised adaptation (LS-LLaMA), including ACEGPT, outperformed zero-shot and few-shot in-context learning approaches, proving more reliable without requiring prompt engineering and despite having as few as 74 training examples. This underscores the importance of fine-tuning models on specialised Arabic datasets.

Although fine-tuned LLMs showed improvements, they were still outperformed by AraBERT, showing the strength of BERT-based models when training sets are small. Prompt engineering was crucial for enhancing model outputs in zero- and few-shot setups, particularly for complex questions. Performance variability across courses and questions indicates the need for more adaptable models. Future research could investigate multitask learning and domain adaptation to provide additional training when few answer examples are available for fine-tuning.

# 9 Limitations and Ethical Considerations

A key limitation of this study is that we were unable to use some larger versions of the models, such as Llama-13B and Jais-70B, due to their high computational requirements. This challenge is compounded by the fact that processing Arabic text demands more computational resources compared to English. Furthermore, given the rapid evolution of large language models and the frequent release of new versions, our study was unable to include the most recent models, such as ChatGPT-4o1 and Llama3, which were released after our experiments were conducted. However, we aimed to draw general findings from the most relevant models available at the time of the study, while also acknowledging the ongoing challenges LLMs face in handling the complexities of the Arabic language. We plan to incorporate newer models in future work.

Ethical considerations are an important aspect of Automated Essay Scoring (AES), particularly regarding the potential for scoring errors. Although AES can help human graders improve consistency and reduce errors, it is essential to implement safeguards to address inaccuracies. Errors in scoring may result in unfair outcomes for students, potentially affecting their academic progress. To mitigate this risk, organizations that use AES should provide transparent feedback mechanisms and clear appeal processes for correcting mistakes.

# References

[1] Yang, K., Raković, M., Li, Y., Guan, Q., Gašević, D., Chen, G.: Unveiling the tapestry of automated essay scoring: A comprehensive investigation of accuracy, fairness, and generalizability. In: Proceedings of the AAAI Conference on Artificial Intelligence, vol. 38, pp. 22466–22474 (2024)

[2] Silveira, I.C., Barbosa, A., Mauá, D.D.: A new benchmark for automatic essay scoring in Portuguese. In: Proceedings of the 16th International Conference on Computational Processing of Portuguese (2024). https://github.com/kamel-usp/aes_enem

[3] Franci, Y., Franci, Y.A.: Enhancing Automated Essay Evaluation: The Impact of Advanced Generative Pre-trained Transformers on Educational Feedback (2023). https://doi.org/10.13140/RG.2.2.13615.51365/1 . https://www.researchgate.net/publication/376271077

[4] Liu, Y., Han, T., Ma, S., Zhang, J., Yang, Y., Tian, J., He, H., Li, A., He, M., Liu, Z., Wu, Z., Zhao, L., Zhu, D., Li, X., Qiang, N., Shen, D., Liu, T., Ge, B.: Summary of ChatGPT-related research and perspective towards the future of Large Language Models. Meta-Radiology **1**, 100017 (2023) https://doi.org/10.1016/j.metrad.2023.100017

[5] Ahmed, T., Devanbu, P.: Few-shot training LLMs for project-specific code-summarization. In: Proceedings of the 37th IEEE/ACM International Conference on Automated Software Engineering, pp. 1–5 (2022)

[6] Xiao, C., Ma, W., Xu, S.X., Zhang, K., Wang, Y., Fu, Q.: From automation to augmentation: Large Language Models elevating essay scoring landscape. arXiv preprint arXiv:2401.06431 (2024)

[7] Mansour, W.A., Albatarni, S., Eltanbouly, S., Elsayed, T.: Can Large Language Models automatically score proficiency of written essays? In: Calzolari, N., Kan, M.-Y., Hoste, V., Lenci, A., Sakti, S., Xue, N. (eds.) Proceedings of the 2024 Joint International Conference on Computational Linguistics, Language Resources and Evaluation (LREC-COLING 2024), pp. 2777–2786. ELRA and ICCL, Torino, Italia (2024). https://aclanthology.org/2024.lrec-main.247

[8] Ghazawi, R., Simpson, E.: Automated essay scoring in Arabic: a dataset and analysis of a Bert-based system. arXiv preprint arXiv:2407.11212 (2024)

[9] Page, E.B.: The imminence of... grading essays by computer. The Phi Delta Kappan **47**(5), 238–243 (1966)

[10] Yannakoudakis, H., Briscoe, T., Medlock, B.: A new dataset and method for automatically grading ESOL texts. In: Proceedings of the 49th Annual Meeting of the Association for Computational Linguistics: Human Language Technologies,

pp. 180–189 (2011)

[11] Chen, H., He, B.: Automated essay scoring by maximizing human-machine agreement. Association for Computational Linguistics, 1741–1752 (2013)

[12] Phandi, P., Chai, K.M.A., Ng, H.T.: Flexible domain adaptation for automated essay scoring using correlated linear regression. In: Proceedings of the 2015 Conference on Empirical Methods in Natural Language Processing, pp. 431–439 (2015)

[13] Dong, F., Zhang, Y., Yang, J.: Attention-based recurrent convolutional neural network for automatic essay scoring. In: Levy, R., Specia, L. (eds.) Proceedings of the 21st Conference on Computational Natural Language Learning (CoNLL 2017), pp. 153–162. Association for Computational Linguistics, Vancouver, Canada (2017). https://doi.org/10.18653/v1/K17-1017 . https://aclanthology.org/K17-1017

[14] Li, X., Yang, H., Hu, S., Geng, J., Lin, K., Li, Y.: Enhanced hybrid neural network for automated essay scoring. Expert Systems **39**(10), 13068 (2022)

[15] Chen, M., Li, X.: Relevance-based automated essay scoring via hierarchical recurrent model. In: 2018 International Conference on Asian Language Processing (IALP), pp. 378–383 (2018). IEEE

[16] Wang, Y., Wang, C., Li, R., Lin, H.: On the use of BERT for automated essay scoring: Joint learning of multi-scale essay representation. arXiv preprint arXiv:2205.03835 (2022)

[17] Yang, R., Cao, J., Wen, Z., Wu, Y., He, X.: Enhancing automated essay scoring performance via fine-tuning pre-trained language models with combination of regression and ranking. In: Findings of the Association for Computational Linguistics: EMNLP 2020, pp. 1560–1569 (2020)

[18] Machhout, R.A., Zribi, C.B.O., Bouzid, S.M.: Arabic automatic essay scoring systems: An overview study. In: International Conference on Intelligent Systems Design and Applications, pp. 1164–1176 (2021). Springer

[19] Alghamdi, M., Alkanhal, M., Al-Badrashiny, M., Al-Qabbany, A., Areshey, A., Alharbi, A.: A hybrid automatic scoring system for Arabic essays. AI Communications **27**(2), 103–111 (2014) https://doi.org/10.3233/AIC-130586

[20] Al-Shalabi, E.F.: An automated system for essay scoring of online exams in Arabic based on stemming techniques and levenshtein edit operations. International Journal of Computer Science Issues **13**(5), 45–50 (2016)

[21] Alqahtani, A., Alsaif, A.: Automated Arabic essay evaluation. In: Proceedings of the 17th International Conference on Natural Language Processing (ICON),

pp. 181–190. NLP Association of India (NLPAI), Indian Institute of Technology Patna, Patna, India (2020). https://aclanthology.org/2020.icon-main.24

[22] Alqahtani, A., Alsaif, A.: Automatic Evaluation for Arabic Essays: A Rule-Based System. 2019 IEEE international symposium on signal processing and information technology (ISSPIT) (2019)

[23] Al-Shargabi, B., Alzyadat, R., Hamad, F.: AEGD: Arabic essay grading dataset for machine learning. Journal of Theoretical and Applied Information Technology (2021)

[24] Gaheen, M.M., ElEraky, R.M., Ewees, A.A.: Automated students Arabic essay scoring using trained neural network by E-Jaya optimization to support personalized system of instruction. Education and Information Technologies **26**(1), 1165–1181 (2021) https://doi.org/10.1007/s10639-020-10300-6

[25] Ramesh, D., Sanampudi, S.K.: An automated essay scoring systems: a systematic literature review. Artificial Intelligence Review **55**(3), 2495–2527 (2022)

[26] Krugmann, J.O., Hartmann, J.: Sentiment analysis in the age of generative AI. Customer Needs and Solutions **11**(1), 3 (2024)

[27] Touvron, H., Martin, L., Stone, K., Albert, P., Almahairi, A., Babaei, Y., Bashlykov, N., Batra, S., Bhargava, P., Bhosale, S., et al.: Llama 2: Open foundation and fine-tuned chat models. arXiv preprint arXiv:2307.09288 (2023)

[28] Touvron, H., Lavril, T., Izacard, G., Martinet, X., Lachaux, M.-A., Lacroix, T., Rozière, B., Goyal, N., Hambro, E., Azhar, F., et al.: Llama: Open and efficient foundation language models. arXiv preprint arXiv:2302.13971 (2023)

[29] Sengupta, N., Sahu, S.K., Jia, B., Katipomu, S., Li, H., Koto, F., Afzal, O.M., Kamboj, S., Pandit, O., Pal, R., et al.: Jais and Jais-Chat: Arabic-centric foundation and instruction-tuned open generative Large Language Models. arXiv preprint arXiv:2308.16149 (2023)

[30] Ammar, A., Koubaa, A., Benjdira, B., Nacar, O., Sibaee, S.: Prediction of Arabic legal rulings using large language models. Electronics **13**(4), 764 (2024)

[31] Tamer, A., Hassan, A.-A., Ali, A., Salah, N., Medhat, W.: Fine tuning of Large Language Models for Arabic Language. In: 2023 20th ACS/IEEE International Conference on Computer Systems and Applications (AICCSA), pp. 1–4 (2023). IEEE

[32] Katuka, G.A., Gain, A., Yu, Y.-Y.: Investigating automatic scoring and feedback using Large Language Models. arXiv preprint arXiv:2405.00602 (2024)

[33] Kasneci, E., Sessler, K., Küchemann, S., Bannert, M., Dementieva, D., Fischer,

F., Gasser, U., Groh, G., Günnemann, S., Hüllermeier, E., Krusche, S., Kutyniok, G., Michaeli, T., Nerdel, C., Pfeffer, J., Poquet, O., Sailer, M., Schmidt, A., Seidel, T., Stadler, M., Weller, J., Kuhn, J., Kasneci, G.: ChatGPT for good? on opportunities and challenges of Large Language Models for education. Learning and Individual Differences **103**, 102274 (2023) https://doi.org/10.1016/j.lindif.2023.102274

[34] Latif, E., Zhai, X.: Fine-tuning ChatGPT for automatic scoring. Computers and Education: Artificial Intelligence **6**, 100210 (2024)

[35] Henkel, O., Hills, L., Roberts, B., McGrane, J.: Can llms grade open response reading comprehension questions? an empirical study using the roars dataset. International Journal of Artificial Intelligence in Education, 1–26 (2024)

[36] Henkel, O., Hills, L., Boxer, A., Roberts, B., Levonian, Z.: Can large language models make the grade? an empirical study evaluating LLMs' ability to mark short answer questions in k-12 education. In: Proceedings of the Eleventh ACM Conference on Learning@ Scale, pp. 300–304 (2024)

[37] Rasul, T., Nair, S., Kalendra, D., Robin, M., Oliveira Santini, F., Ladeira, W.J., Sun, M., Day, I., Rather, R.A., Heathcote, L.: The role of ChatGPT in higher education: Benefits, challenges, and future research directions. Journal of Applied Learning and Teaching **6**(1) (2023)

[38] Koto, F., Li, H., Shatnawi, S., Doughman, J., Sadallah, A.B., Alraeesi, A., Almubarak, K., Alyafeai, Z., Sengupta, N., Shehata, S., et al.: ArabicMMLU: Assessing massive multitask language understanding in Arabic. arXiv preprint arXiv:2402.12840 (2024)

[39] Achiam, J., Adler, S., Agarwal, S., Ahmad, L., Akkaya, I., Aleman, F.L., Almeida, D., Altenschmidt, J., Altman, S., Anadkat, S., et al.: GPT-4 technical report. arXiv preprint arXiv:2303.08774 (2023)

[40] Ghosn, Y., El Sardouk, O., Jabbour, Y., Jrad, M., Hussein Kamareddine, M., Abbas, N., Saade, C., Abi Ghanem, A.: ChatGPT 4 versus ChatGPT 3.5 on the final FRCR part a sample questions. assessing performance and accuracy of explanations. medRxiv, 2023–09 (2023)

[41] Massey, P.A., Montgomery, C., Zhang, A.S.: Comparison of ChatGPT–3.5, ChatGPT-4, and orthopaedic resident performance on orthopaedic assessment examinations. JAAOS-Journal of the American Academy of Orthopaedic Surgeons **31**(23), 1173–1179 (2023)

[42] Plevris, V., Papazafeiropoulos, G., Jiménez Rios, A.: Chatbots put to the test in math and logic problems: a comparison and assessment of ChatGPT-3.5, ChatGPT-4, and Google Bard. AI **4**(4), 949–969 (2023)

[43] Üstün, A., Aryabumi, V., Yong, Z.-X., Ko, W.-Y., D'souza, D., Onilude, G., Bhandari, N., Singh, S., Ooi, H.-L., Kayid, A., et al.: Aya model: An instruction finetuned open-access multilingual language model. arXiv preprint arXiv:2402.07827 (2024)

[44] Alyafeai, Z., Masoud, M., Ghaleb, M., Al-shaibani, M.S.: Masader: Metadata sourcing for Arabic text and speech data resources. In: Calzolari, N., Béchet, F., Blache, P., Choukri, K., Cieri, C., Declerck, T., Goggi, S., Isahara, H., Maegaard, B., Mariani, J., Mazo, H., Odijk, J., Piperidis, S. (eds.) Proceedings of the Thirteenth Language Resources and Evaluation Conference, pp. 6340–6351. European Language Resources Association, Marseille, France (2022). https://aclanthology.org/2022.lrec-1.681

[45] Muennighoff, N., Wang, T., Sutawika, L., Roberts, A., Biderman, S., Le Scao, T., Bari, M.S., Shen, S., Yong, Z.X., Schoelkopf, H., Tang, X., Radev, D., Aji, A.F., Almubarak, K., Albanie, S., Alyafeai, Z., Webson, A., Raff, E., Raffel, C.: Crosslingual generalization through multitask finetuning. In: Rogers, A., Boyd-Graber, J., Okazaki, N. (eds.) Proceedings of the 61st Annual Meeting of the Association for Computational Linguistics (Volume 1: Long Papers), pp. 15991–16111. Association for Computational Linguistics, Toronto, Canada (2023). https://doi.org/10.18653/v1/2023.acl-long.891 . https://aclanthology.org/2023.acl-long.891

[46] Huang, H., Yu, F., Zhu, J., Sun, X., Cheng, H., Dingjie, S., Chen, Z., Alharthi, M., An, B., He, J., Liu, Z., Chen, J., Li, J., Wang, B., Zhang, L., Sun, R., Wan, X., Li, H., Xu, J.: AceGPT, localizing large language models in Arabic. Human Language Technologies (Volume 1: Long Papers), 8139–8163 (2024) https://doi.org/10.18653/v1/2024.naacl-long.450

[47] Demidova, A., Atwany, H., Rabih, N., Sha'ban, S.: Arabic train at NADI 2024 shared task: LLMs' ability to translate Arabic dialects into Modern Standard Arabic. In: Habash, N., Bouamor, H., Eskander, R., Tomeh, N., Abu Farha, I., Abdelali, A., Touileb, S., Hamed, I., Onaizan, Y., Alhafni, B., Antoun, W., Khalifa, S., Haddad, H., Zitouni, I., AlKhamissi, B., Almatham, R., Mrini, K. (eds.) Proceedings of The Second Arabic Natural Language Processing Conference, pp. 729–734. Association for Computational Linguistics, Bangkok, Thailand (2024). https://doi.org/10.18653/v1/2024.arabicnlp-1.80 . https://aclanthology.org/2024.arabicnlp-1.80

[48] Mathias, S., Bhattacharyya, P.: ASAP++: Enriching the ASAP automated essay grading dataset with essay attribute scores. In: Proceedings of the Eleventh International Conference on Language Resources and Evaluation (LREC 2018) (2018)

[49] Bari, M.S., Alnumay, Y., Alzahrani, N.A., Alotaibi, N.M., Alyahya, H.A., Al-Rashed, S., Mirza, F.A., Alsubaie, S.Z., Alahmed, H.A., Alabduljabbar, G., et

al.: ALLaM: Large Language Models for Arabic and English. arXiv preprint arXiv:2407.15390 (2024)

[50] Petrov, A., La Malfa, E., Torr, P., Bibi, A.: Language model tokenizers introduce unfairness between languages. Advances in Neural Information Processing Systems **36** (2024)

[51] Kwon, S., Bhatia, G., Abdul-Mageed, M., *et al.*: Beyond English: Evaluating LLMs for Arabic grammatical error correction. In: Proceedings of ArabicNLP 2023, pp. 101–119 (2023)

[52] Patil, R., Gudivada, V.: A review of current trends, techniques, and challenges in Large Language Models (LLMs). Applied Sciences **14**(5), 2074 (2024)

[53] Edwards, A., Camacho-Collados, J.: Language models for text classification: Is in-context learning enough? arXiv preprint arXiv:2403.17661 (2024)

[54] Chamieh, I., Zesch, T., Giebermann, K.: LLMs in short answer scoring: Limitations and promise of zero-shot and few-shot approaches. In: Kochmar, E., Bexte, M., Burstein, J., Horbach, A., Laarmann-Quante, R., Tack, A., Yaneva, V., Yuan, Z. (eds.) Proceedings of the 19th Workshop on Innovative Use of NLP for Building Educational Applications (BEA 2024), pp. 309–315. Association for Computational Linguistics, Mexico City, Mexico (2024). https://aclanthology.org/2024.bea-1.25

[55] Gema, A., Daines, L., Minervini, P., Alex, B.: Parameter-efficient fine-tuning of Llama for the clinical domain. arXiv preprint arXiv:2307.03042 (2023)

[56] Li, Z., Li, X., Liu, Y., Xie, H., Li, J., Wang, F.-l., Li, Q., Zhong, X.: Label supervised Llama finetuning. arXiv preprint arXiv:2310.01208 (2023)

[57] Geng, X., Liu, H.: OpenLLaMA: An Open Reproduction of LLaMA (2023). https://github.com/openlm-research/open_llama

[58] Kocoń, J., Cichecki, I., Kaszyca, O., Kochanek, M., Szydło, D., Baran, J., Bielaniewicz, J., Gruza, M., Janz, A., Kanclerz, K., *et al.*: ChatGPT: Jack of all trades, master of none. Information Fusion **99**, 101861 (2023)

# Appendix A  Tables

Table A1: Performance of Models in Predicting Scores Within One Mark (±1 score) of Actual Score.

| Approach | Fine-Tuning | Few-shot | | | | Zero-shot | |
|---|---|---|---|---|---|---|---|
| Dataset / Model | Llama2 (Instruction) ±1 score | ACEGPT ±1 score | Aya ±1 score | ChatGPT-4 ±1 score | Jais ±1 score | Jais ±1 score | ChatGPT-4 |
| **Entire Dataset** | 41.03% | 72.15% | 57.97% | 83.44% | 52.06% | 49.24% | 72.06% |
| **Intro to Info Sci** | 58.82% | 67.50% | 60.00% | 77.04% | 60.82% | 45.58 | 66.39% |
| **Management Info Sys** | 90.90% | 74.00% | 74.19% | 91.30% | 37.5% | 31.81 | 88.40 % |
| **Environmental Chemistry** | 59.57% | 82.00% | 82.85% | 90.38% | 50.00% | 51.51% | 76.92% |
| **Biotechnology** | 84.37% | 68.42% | 63.63% | 77.08% | 53.65% | 54.83% | 81.25% |
| **Average of Questions (Q1-Q12)** | 73.99% | 69.58% | 61.31% | 79.55% | 60.45% | 41.65% | 83.03% |

# Appendix B  Figures

**Prompt Instructions**

### Instruction:
Mark the following essay based on the provided instructions and details: Assign one of the following six scores: 0, 1, 2, 3, 4, or 5 (where 0 = Very Poor and 5 = Excellent). Provide only the final numerical score without any additional comments or explanations.

# The Question:
أشرح باستفاضة دور زيادة التخصصات الدقيقة وتزايد الموضوعات في التأثير على ثورة انفجار المعلومات ؟

# Criteria for Evaluation:
1. قدرة الطالب على شرح أسباب تزايد التخصصات والموضوعات. (2 درجتان)
2. قدرة الطالب على دور وتأثير زيادة التخصصية في ثورة المعلومات. (2 درجتان)
3. قدرة الطالب على ربط أسباب ظهور علوم حديثة مع انفجار المعلومات. (1 درجة)

# Standard Answer:
يعد التخصص الدقيق وتزايد الموضوعات أحد عناصر ثورة المعلومات وهذا بسبب التوجه العالمي المتزايد في...

### Input:
كلما زاد عدد الباحثون وكبر حجم الانتاج الفكري كان سبب في قلت فاعلية الدوريات التي تعمل على تغطية ...

# Response: 2
### End

**Figure B1**: An Example of fine-tuning experimentation with Llama instruction, illustrating the prompt engineering used.